\documentclass{article}
\usepackage[T1]{fontenc}
\usepackage{spconf,amsmath,amssymb,graphicx}
\usepackage{booktabs,multirow,pifont,balance}
\usepackage[hidelinks]{hyperref}

\newcommand{\best}[1]{\textbf{#1}}
\newcommand{\second}[1]{\underline{#1}}
\title{3DGS-SC: A Controlled Static Screen-Content Benchmark for 3D Gaussian Splatting}

\name{Shicheng Cai$^{1}$ \qquad Hao Zhang$^{1}$ \qquad Dong Dai$^{1}$ \qquad Xuerui Ma$^{2}$ \qquad Ying Hu$^{2}$ \qquad Tao Zhang$^{1}$}

\address{$^{1}$ Central South University, Changsha, China \\
$^{2}$ Hunan Malanshan Audio and Video Laboratory, China \\
\texttt{244511028@csu.edu.cn}}
\begin{document}
\ninept
\raggedbottom
\maketitle
\begin{abstract}
Does high image fidelity imply readable screen content in 3D Gaussian Splatting (3DGS)? We introduce 3DGS-SC, a controlled static screen-content dataset and benchmark for examining this mismatch. Ten procedural scenes provide fixed multi-view splits, exact cameras, screen masks, text boxes, and transcripts. The protocol separates whole-image fidelity, screen-region fidelity, OCR readability, and edge preservation. In the reported five-method comparison, LightGaussian exceeds Mip-Splatting in screen PSNR by only 0.08 dB, yet trails it in OCR accuracy by 12.7 percentage points. Across all ten method pairs, screen-PSNR and OCR orderings disagree in seven cases. These aggregate results expose a method-selection failure of fidelity-only evaluation. Complementing prior text-aware 3DGS research, 3DGS-SC targets controlled monitor interfaces and exact annotations; scene-wise robustness and acquisition effects remain open validation questions.

\end{abstract}

\section{Introduction}
3D Gaussian Splatting (3DGS) enables efficient novel view synthesis with competitive image quality~\cite{kerbl2023gaussian,yu2024mipsplatting}. Natural-scene evaluation commonly emphasizes PSNR and SSIM~\cite{mildenhall2020nerf,jensen2014dtu,knapitsch2017tanks,barron2022mipnerf360}. For screen content, however, the information-bearing units are characters, table borders, and interface separators. A small image error can alter a character or break a thin line without strongly changing an average fidelity score.

We study a specific evaluation problem: \emph{does choosing a 3DGS method by image fidelity also select a method that preserves screen readability?} This is distinct from merely observing blurred text. It tests whether the benchmark's selection criterion agrees with the information a screen must convey. Masking out the background may remove score dilution, but does not by itself make pixel similarity sensitive to character identity.

Text degradation in 3DGS is not an unstudied phenomenon. STRinGS introduces text refinement, OCR-CER evaluation, and the STRinGS-360 dataset~\cite{raundhal2026strings}. Our focus is a controlled static monitor setting with procedural interfaces, fixed cameras, and exact screen and text annotations. These controls are intended to separate rendering failures from glare, sensor noise, and physical display effects.

Our contributions are: (i) a dataset and annotation design for static screen-content NVS; (ii) a protocol that separates global fidelity, screen fidelity, readability, and structure; and (iii) a quantitative analysis of their disagreement across five reported baselines. The strongest counterexample is not a background effect: methods only 0.08 dB apart in screen PSNR differ by 12.7 percentage points in OCR accuracy. We interpret this as aggregate evidence, without claiming scene-wise significance or a causal explanation.

\begin{figure}[t]
\centering
\includegraphics[width=\columnwidth]{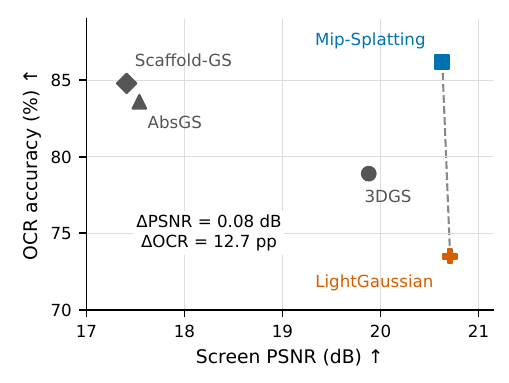}
\caption{Screen fidelity versus readability on 3DGS-SC. Each point is one method's reported aggregate in Table~\ref{tab:overall_results}. LightGaussian and Mip-Splatting differ by only 0.08 dB in screen PSNR but by 12.7 percentage points in OCR accuracy. The dashed segment connects this comparison, not a fitted trend. Axes show the observed score range; per-scene uncertainty is unavailable.}
\label{fig:teaser}
\end{figure}

\section{Related Works}
\label{sec:related-works}
\subsection{3D Gaussian Splatting and Its Variants}
Modern view-conditioned scene reconstruction was established by neural radiance fields ~\cite{mildenhall2020nerf}, with efficient explicit variants such as Plenoxels and Instant-NGP broadening the representation space~\cite{fridovichkeil2022plenoxels,mueller2022instantngp}. 3D Gaussian Splatting (3DGS) later introduced an explicit anisotropic Gaussian representation with differentiable splat rendering and real-time performance ~\cite{kerbl2023gaussian}. Subsequent variants mainly improve one of three properties: anti-aliasing, fine-detail recovery, or efficiency. Mip-Splatting reduces scale-dependent artifacts ~\cite{yu2024mipsplatting}; AbsGS improves small-structure reconstruction ~\cite{ye2024absgs}; Scaffold-GS introduces a structured view-adaptive parameterization ~\cite{lu2024scaffoldgs}; and LightGaussian targets compact deployment~\cite{fan2024lightgaussian}. These methods span representation, sampling, and efficiency choices relevant to fine screen structures.

\subsection{Novel View Synthesis Benchmarks}
Mainstream NVS and 3DGS benchmarks still focus on natural scenes. NeRF-Synthetic ~\cite{mildenhall2020nerf}, DTU~\cite{jensen2014dtu,aanaes2016dtu}, Tanks and Temples~\cite{knapitsch2017tanks}, ScanNet++~\cite{yeshwanth2023scannetpp}, DL3DV-10K~\cite{ling2024dl3dv10k}, and MegaScenes~\cite{tung2024megascenes} are valuable for measuring photorealistic reconstruction. Their standard whole-image evaluations do not directly measure screen text readability. Text-rich captured scenes in STRinGS-360 provide a closer reference~\cite{raundhal2026strings}; our procedural monitor setting instead emphasizes exact annotations and controlled interface layouts.

\subsection{Screen Content Coding and Quality Evaluation}
The screen-content literature has long recognized that computer-generated layouts exhibit statistics distinct from natural imagery. Screen-content image quality studies show that generic perceptual metrics are often insufficient without region-aware, structure-aware, or text-aware analysis~\cite{yang2014subjective,gu2015siqm,ni2016edgemodel,ni2017scid,wang2018rrsci,chen2020frsciqa,dong2020screenqa}. Our goal is different from building a 2D screen codec or a no-reference image-quality metric. Instead, we bring screen-aware evaluation into the static multi-view 3DGS setting through a controlled dataset, exact annotations, and a reproducible benchmark protocol.

\section{The 3DGS-SC Benchmark}
\label{sec:method}

\subsection{Design Principles}
\label{subsec:data_collection}
3DGS-SC targets static multi-view screen-content scenes in which the screen is the information-bearing object of interest. The benchmark is designed around three principles. First, controlled difficulty: scene variation is introduced through explicit factors including scene family, monitor layout, and camera pose, while unrelated clutter and capture artifacts are removed. Second, reproducibility: all scenes are generated from scripts, released in a NeRF-style directory structure, and paired with deterministic train/validation/test splits. Third, screen-aware evaluation: each rendered view is accompanied by the annotations required to score both conventional image fidelity and screen-content-specific usability.

\begin{figure*}[t]
    \centering
    \includegraphics[width=\textwidth]{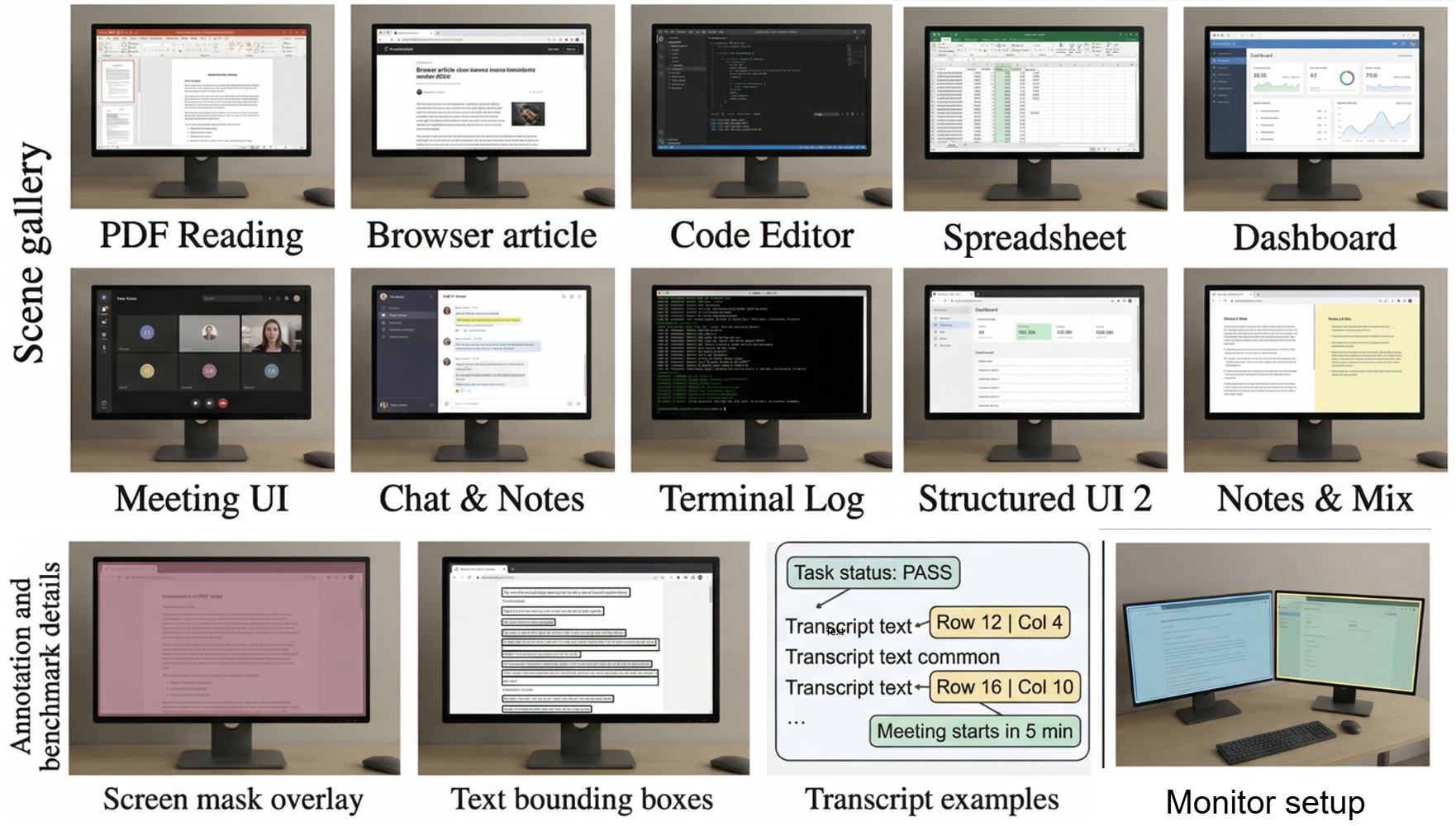}
    \caption{Representative scenes and annotations in 3DGS-SC. The top two rows show diverse screen-content scenes, including reading, code editing, spreadsheet, dashboard, meeting, and chat layouts. The bottom row visualizes benchmark annotations, including screen masks, text bounding boxes, aligned transcripts, and a monitor-setup example, demonstrating the screen-aware annotation package released with the benchmark.}
    \label{fig:showcase_annotations}
\end{figure*}

\begin{table*}[t]
\centering
\caption{Scope of representative evaluation datasets. STRinGS-360 already studies text-rich 3DGS; the proposed distinction is controlled screen content and exact procedural annotations, not the first use of OCR for 3DGS.}
\label{tab:dataset_comparison}
\small
\setlength{\tabcolsep}{4pt}
\begin{tabular}{llll}
\toprule
Dataset & Domain & Acquisition & Evaluation emphasis \\
\midrule
NeRF-Synthetic~\cite{mildenhall2020nerf} & Objects & Synthetic & Multi-view image fidelity \\
Mip-NeRF 360~\cite{barron2022mipnerf360} & Natural scenes & Captured & Multi-view image fidelity \\
SCID~\cite{ni2017scid} & Screen images & 2D image database & Screen image quality \\
STRinGS-360~\cite{raundhal2026strings} & Text-rich objects & Captured & Text-aware 3DGS / OCR-CER \\
\textbf{3DGS-SC} & Static monitor interfaces & Procedural & Screen fidelity, OCR, edges \\
\bottomrule
\end{tabular}
\end{table*}

\begin{table*}[t]
\centering
\caption{Overall benchmark results on 3DGS-SC. Best values are bold; second-best values are underlined (including ties). 
OCR accuracy and CER are expressed as percentages; PSNR is in dB. Values are inherited from the manuscript aggregates. 
The table shows that global fidelity and SC-oriented quality are not equivalent.}
\label{tab:overall_results}
\setlength{\tabcolsep}{5pt}
\renewcommand{\arraystretch}{1.10}
\begin{tabular}{lcccccc}
\toprule
\multirow{2}{*}{Method} 
& \multicolumn{2}{c}{Global Fidelity} 
& \multicolumn{3}{c}{SC-aware Quality} 
& \multicolumn{1}{c}{Structure} \\
\cmidrule(lr){2-3}\cmidrule(lr){4-6}\cmidrule(lr){7-7}
& PSNR$\uparrow$ & SSIM$\uparrow$ 
& PSNR$_{\text{screen}}\uparrow$ & OCR Acc.$\uparrow$ & CER$\downarrow$ 
& Edge-F1$\uparrow$  \\
\midrule
3DGS          & 11.67 & 0.79 & 19.88 & 78.9 & 12.8 & 0.781 \\
Mip-Splatting & 11.18 & \second{0.84} & \second{20.63} & \best{86.2} & \best{7.4} & \best{0.842} \\
AbsGS         & \second{13.41} & 0.74 & 17.54 & 83.6 & 8.7 & \second{0.834} \\
Scaffold-GS   & \best{28.96} & \second{0.84} & 17.41 & \second{84.8} & \second{8.1} & 0.828 \\
LightGaussian & 11.94 & \best{0.92} & \best{20.71} & 73.5 & 17.2 & 0.744 \\
\bottomrule
\end{tabular}
\end{table*}

\subsection{Dataset Construction}
\label{subsec:annotations}
We construct the benchmark from ten procedurally generated screen-content scenes covering document review, browser article, PDF reading, code editor, spreadsheet, terminal log, dense slide, dashboard, meeting/share interface, and chat/notes interface. Because all assets are computer-generated, layout and text content are exactly known at generation time. This allows us to release screen masks, text bounding boxes, and aligned transcripts without manual annotation.
Each asset is instantiated in a monitor-centric Blender scene containing only the necessary display geometry, including the monitor bezel, back shell, stem, and base. The screen is modeled as an emissive surface under neutral illumination. We consider four display configurations: Single-Front, Single-Tilted, Dual-Angled, and Dual-Asymmetric. For dual-monitor scenes, the original screen asset is partitioned across displays rather than duplicated, preserving semantic continuity in wide-workspace layouts.
Views are rendered under a fixed monitor-centered camera protocol parameterized by azimuth, elevation, and distance. All valid views satisfy a whole-screen visibility constraint so that evaluation emphasizes viewpoint-induced distortion rather than partial cropping. Each scene provides 20 training views, 20 validation views, and 20 test views, with held-out test poses biased toward more oblique viewpoints.

\subsection{Screen-Aware Annotations and Release Package}
\label{subsec:release}
Each rendered view is paired with a mandatory annotation package. Camera intrinsics, extrinsics, and split labels are stored in transforms\_train.json, transforms\_val.json, and transforms\_test.json. A binary screen mask identifies the visible screen region at the original image resolution. Axis-aligned text boxes are expressed directly in final-image pixel coordinates, and each box is paired one-to-one with the exact rendered transcript, including numerals, punctuation, and letter case.
The release specification comprises RGB(A) images, camera files, screen masks, aligned text annotations, an evaluation script, baseline configurations, generation scripts, seeds, and result templates. The protocol first averages per-view scores within each scene and then macro-averages across scenes. This prevents scenes with more views from dominating the aggregate; text instances are aggregated within each view. The present analysis uses the reported method aggregates rather than recomputing them from per-view records.

\subsection{SC-Oriented Evaluation Protocol}
\label{subsec:evaluation_protocol}

The benchmark separates four quality axes. Efficiency is an optional reporting axis, not evidence for the present quality comparison.

\subsubsection{Whole-Image Fidelity}
\label{subsubsec:whole_image_metrics}

We report PSNR and SSIM on complete images; LPIPS is a protocol extension not included in the current result table. These metrics maintain compatibility with standard 3DGS evaluation practice~\cite{kerbl2023gaussian,yu2024mipsplatting,barron2022mipnerf360}.

\subsubsection{Screen-Region Quality}
\label{subsubsec:screen_metrics}

Because SC quality is primarily determined by the rendered screen area, we additionally compute region-restricted metrics within the screen mask. Let $\Omega_s$ denote the screen region. For a metric $\mathcal{M}$, its screen-aware version is defined as
\begin{equation}
\mathcal{M}_{\mathrm{screen}} = \mathcal{M}(\hat{I}|_{\Omega_s}, I|_{\Omega_s}),
\end{equation}
where $\hat{I}$ and $I$ denote the predicted and reference images, respectively. The current comparison reports screen-region PSNR. Masked SSIM and LPIPS require explicit window/crop handling and are left outside this analysis.

\subsubsection{OCR-Based Text Readability}
\label{subsubsec:ocr_metrics}

We evaluate released text regions using a fixed Tesseract-style OCR backend. Each rendered crop undergoes the released preprocessing pipeline and is compared with its aligned transcript. Exact-match accuracy measures whether an instance is recovered correctly; character error rate (CER) measures partial corruption. The scorer fixes preprocessing and OCR configuration across methods. OCR is a standardized readability probe, interpreted jointly with screen-region fidelity and Edge-F1 rather than as the sole measure of usability.

\subsubsection{Edge-Sensitive Structure Preservation}
\label{subsubsec:edge_metrics}

Text strokes, table borders, and thin UI separators are particularly sensitive to blur and aliasing, as also emphasized in screen-content quality assessment research~\cite{gu2015siqm,ni2016edgemodel}. We therefore compute an edge-F1 score inside the screen mask. Reference and rendered edge maps are extracted with the same Canny parameters, and edge pixels are matched with a two-pixel tolerance before computing precision, recall, and their harmonic mean. This score complements OCR by measuring whether screen structures remain geometrically stable even when the OCR engine can still guess the correct word.

\subsubsection{Efficiency-Related Measures}
\label{subsubsec:efficiency_metrics}

Model size, Gaussian count, and rendering speed are recommended companion measurements. They are not available in the current table; no measured quality--efficiency trade-off or compression effect is inferred here.

\section{Experiments}
\label{sec:experiments}

\subsection{Experimental Setup}
\label{subsec:exp_setup}

We benchmark five representative 3DGS baselines on 3DGS-SC:  \textbf{3DGS}~\cite{kerbl2023gaussian}, \textbf{Mip-Splatting}~\cite{yu2024mipsplatting}, \textbf{AbsGS}~\cite{ye2024absgs}, \textbf{Scaffold-GS}~\cite{lu2024scaffoldgs}, and \textbf{LightGaussian}~\cite{fan2024lightgaussian}. These methods cover complementary design motivations relevant to screen-content rendering, including canonical Gaussian optimization, anti-aliasing under scale variation, fine-detail recovery, structured representation, and compact deployment.

Following the protocol in Section~\ref{subsec:evaluation_protocol}, we report \textbf{PSNR} and \textbf{SSIM} as whole-image metrics~\cite{kerbl2023gaussian,barron2022mipnerf360}, \textbf{PSNR$_{\text{screen}}$} as a screen-region metric, \textbf{OCR Accuracy and CER} as readability metric, and \textbf{Edge-F1} as a structure-preservation metric~\cite{gu2015siqm,ni2016edgemodel}. The benchmark comparison is summarized in Table~\ref{tab:overall_results}, while Table~\ref{tab:dataset_comparison} positions 3DGS-SC against representative NVS/3DGS and screen-content-related datasets.

\subsection{Overall Results}
\label{subsec:overall_results}
\textbf{Screen fidelity does not determine readability.}
Table~\ref{tab:overall_results} and Fig.~\ref{fig:teaser} provide a direct counterexample. LightGaussian has screen PSNR 20.71 dB and OCR accuracy 73.5\%; Mip-Splatting has 20.63 dB and 86.2\%, respectively. Selecting the former by screen PSNR gains 0.08 dB but loses 12.7 percentage points in OCR accuracy. Its CER is also higher (17.2\% versus 7.4\%) and Edge-F1 lower (0.744 versus 0.842). Thus the discrepancy persists after restricting fidelity to the screen and is corroborated by a structural metric.

\textbf{Why a screen mask is insufficient.}
A screen mask removes background pixels from the fidelity calculation, but the remaining pixel errors are still averaged without recognizing character identity. The observed counterexample shows that region restriction alone does not make PSNR a substitute for OCR. Table~\ref{tab:overall_results} also identifies different leaders: Scaffold-GS for global PSNR, LightGaussian for SSIM and screen PSNR, and Mip-Splatting for readability and edges. These different leaders in Table~\ref{tab:overall_results} motivate evaluating fidelity and readability separately. These are descriptive comparisons of the supplied averages, not new training runs.

\subsection{Quantifying Ranking Disagreement}
\label{subsec:metric_mismatch}
For higher-is-better metric $q$ and OCR accuracy $a$, define
\begin{equation}
D(q,a)=\frac{\sum_{i<j}\mathbf{1}[(q_i-q_j)(a_i-a_j)<0]}{\sum_{i<j}\mathbf{1}[(q_i-q_j)(a_i-a_j)\ne0]}.
\label{eq:discordance}
\end{equation}
Tied pairs are excluded; CER is sign-reversed when used as $q$. The reported table yields disagreement in 5/10 pairs for global PSNR, 5/9 for SSIM, and 7/10 for screen PSNR. CER agrees with OCR on all ten pairs, while Edge-F1 disagrees on one. These counts quantify the metric-dependent selection problem across the five methods; the pairs are not independent observations and do not establish statistical significance.

\subsection{Evidence Scope and Required Controls}
\label{subsec:discussion}
The evidence is limited to aggregate scores on ten synthetic scenes. Per-scene results, repeated runs, and reference-image OCR scores are not available for this analysis. Consequently, neither scene-wise robustness nor the fraction of OCR error caused specifically by reconstruction can be established. A reference OCR control should compare identical crops and transcripts on reference and rendered images. Controlled sweeps of projected character height, camera obliquity, and text density would then test when the mismatch increases; these are validation experiments, not results reported here.

Training budgets, initialization, output resolution, OCR configuration, and scene-to-layout assignments must accompany a reproducible release. In particular, the reported global PSNR spans 11.18--28.96 dB and should be checked against consistent image ranges, masks, and backgrounds. The benchmark excludes physical display acquisition effects. It complements existing text-aware 3DGS work rather than claiming that text degradation itself is newly discovered.

\section{Conclusion}
3DGS-SC frames static screen-content NVS as a joint fidelity and readability evaluation problem. Its controlled dataset design supports exact screen and text annotations. The reported five-method comparison reveals a 12.7-point OCR gap at nearly identical screen PSNR and seven discordant screen-PSNR/OCR orderings among ten method pairs. These findings motivate screen-aware method selection; scene-wise controls and reference OCR calibration are needed to establish robustness and explain the failure conditions.

\clearpage
\balance
\bibliographystyle{IEEEbib}
\bibliography{arxiv}
\end{document}